\pdfoutput=1
\documentclass[10pt,twocolumn,letterpaper,driverfallback=dvipdfmx]{article}

\usepackage{iccv}
\usepackage{times}
\usepackage{epsfig}
\usepackage{amsmath}
\usepackage{amssymb}

\usepackage{microtype}
\usepackage{graphicx}
\usepackage{color}
\usepackage{subfigure}
\usepackage{booktabs} 
\usepackage{bm}
\usepackage{here}
\usepackage{subfigure}
\usepackage[binary-units]{siunitx}
\usepackage{multirow}
\usepackage{algorithmicx}
\usepackage[ruled]{algorithm}
\usepackage[noend]{algpseudocode}
\usepackage{comment}
\usepackage{cite}

\usepackage[pagebackref=true,breaklinks=true,letterpaper=true,colorlinks,bookmarks=false]{hyperref}
\iccvfinalcopy 

\def\iccvPaperID{30} 

\ificcvfinal\fi

\begin{document}

\title{Rethinking Task and Metrics of Instance Segmentation on 3D Point Clouds}

\author{Kosuke Arase${}^{1}$, Yusuke Mukuta${}^{1,2}$, Tatsuya Harada${}^{1,2}$\\
${}^{1}$The University of Tokyo ${}^{2}$RIKEN AIP\\
{\tt\small\{arase,mukuta,harada\}@mi.t.u-tokyo.ac.jp}
}

\maketitle

\begin{abstract}
Instance segmentation on 3D point clouds is one of the most extensively researched areas toward the realization of autonomous cars and robots.
Certain existing studies have split input point clouds into small regions such as \SI{1}{\metre}$\times$\SI{1}{\metre}; one reason for this is that models in the studies cannot consume a large number of points because of the large space complexity.
However, because such small regions occasionally include a very small number of instances belonging to the same class, an evaluation using existing metrics such as mAP is largely affected by the category recognition performance.
To address these problems, we propose a new method with space complexity $\mathcal{O}(N_p)$ such that large regions can be consumed, as well as novel metrics for tasks that are independent of the categories or size of the inputs.
Our method learns a mapping from input point clouds to an embedding space, where the embeddings form clusters for each instance and distinguish instances using these clusters during testing.
Our method achieves state-of-the-art performance using both existing and the proposed metrics. Moreover, we show that our new metric can evaluate the performance of a task without being affected by any other condition.

\end{abstract}

\section{Introduction}
\label{sec:introduction}
3D environment recognition has been extensively researched toward the realization of autonomous cars and robots.
In particular, instance segmentation, the task of not only labeling each point but also distinguishing each instance belonging to the same class, is one of the key tasks to such realization.
Instance segmentation is challenging because the number of instances is not fixed, and thus, methods for categorical classification cannot be directly applied.
Although there are several typical 3D data representations such as voxels, meshes, and point clouds, in this study, we focus on point clouds, which can be obtained directly from depth sensors such as Light Detection and Ranging (LiDAR).

\begin{figure*}[t]
    \centering
    \subfigure[\SI{1}{\metre}$\times$\SI{1}{\metre}] {
        \includegraphics[height=3.5cm]{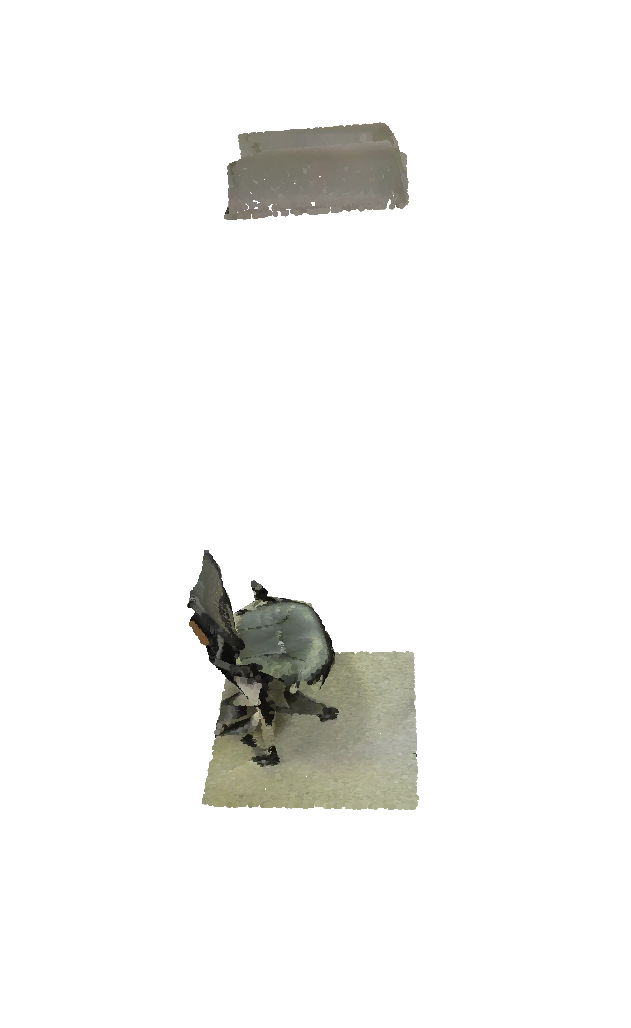}}
    \hspace{10mm}
    \subfigure[\SI{3}{\metre}$\times$\SI{3}{\metre}] {
        \includegraphics[height=3.5cm]{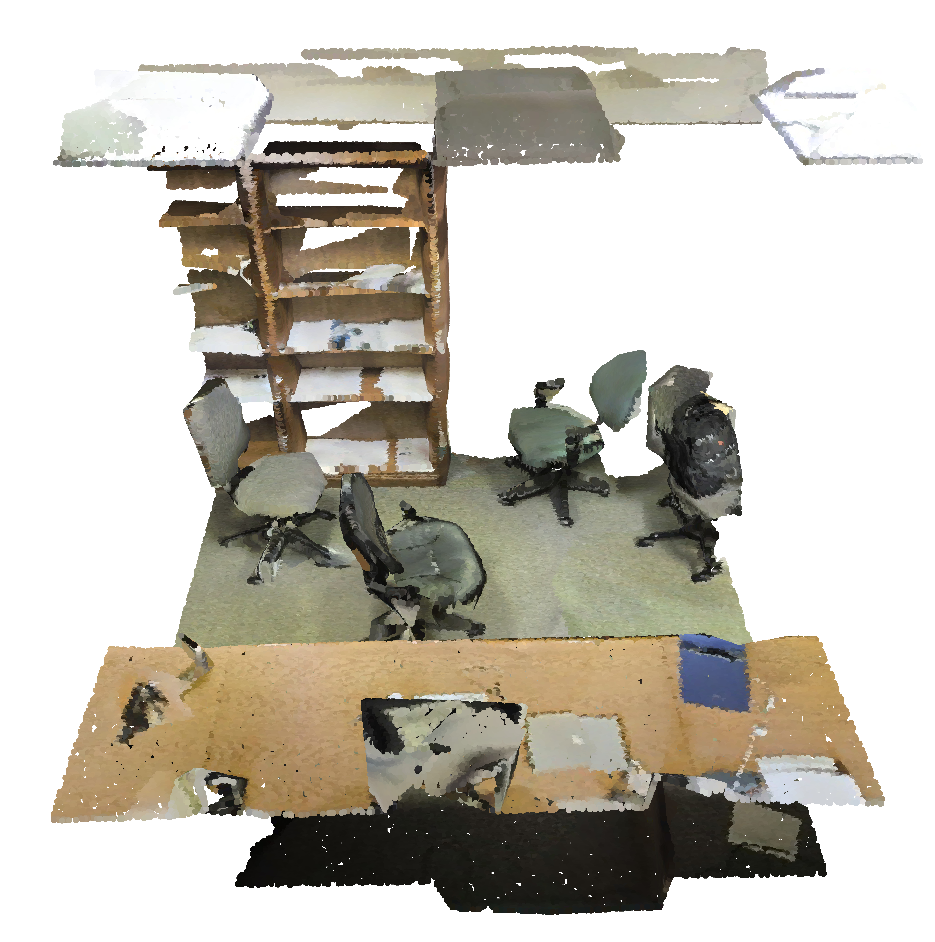}}
    \hspace{10mm}
    \subfigure[\SI{6}{\metre}$\times$\SI{6}{\metre}] {
        \includegraphics[height=3.5cm]{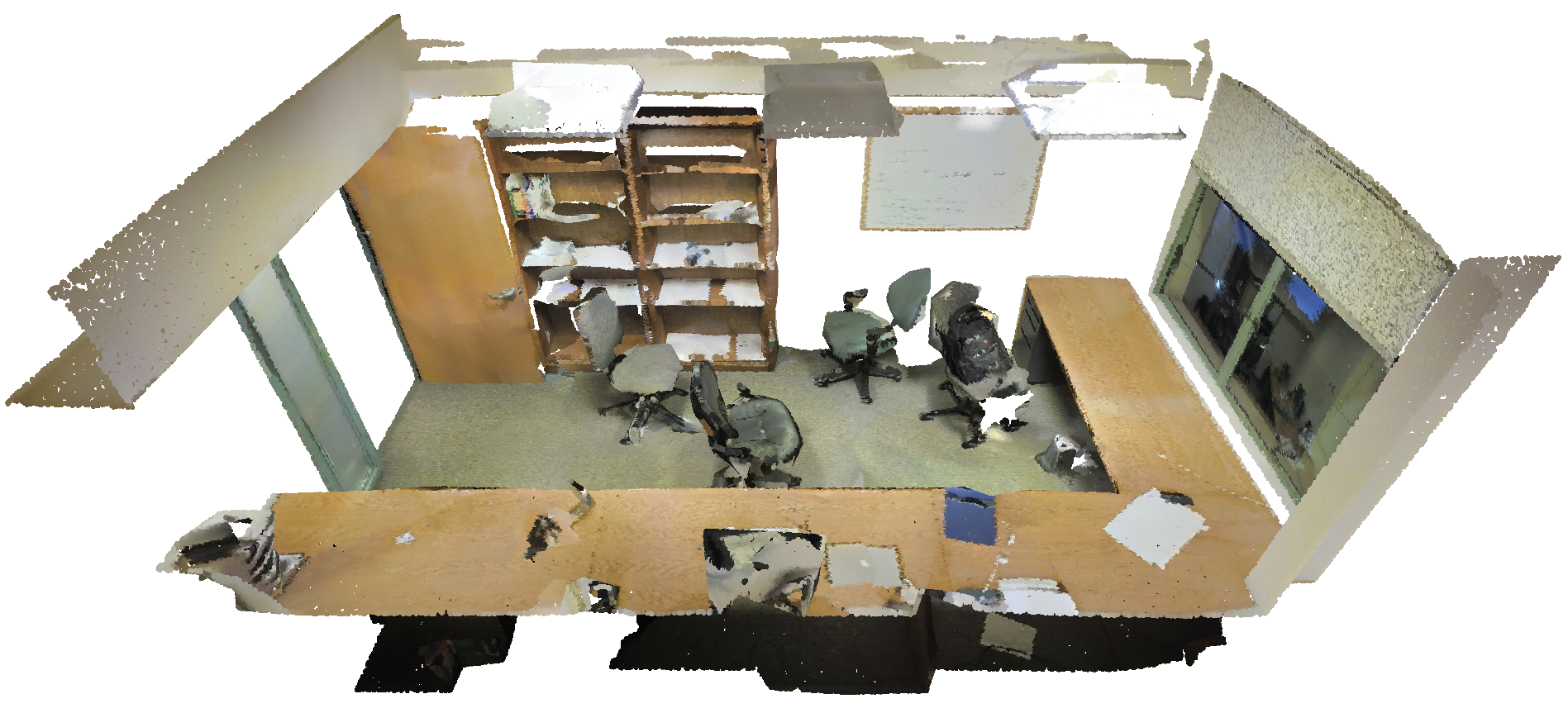}}
    \caption{\label{fig:size_of_grid} Objects within grids for various grid sizes (S3DIS \cite{Armeni2016}, Area 6, Office 29)}
\end{figure*}

The instance segmentation model learns the mapping from each input point to the semantics of the corresponding point.
When evaluating the instance segmentation model, the pairing of a prediction and a ground truth is considered true positive when the intersection over union (IoU) between them is higher than the threshold.
In many cases, semantic segmentation can be solved simultaneously, and thus, the important issue is distinguishing objects in the same category.

There have been many studies on instance segmentation, where the input point clouds have been split into small regions such as \SI{1}{\metre} square \cite{Wang2017,sung2018}; however, conducting evaluations on such small regions is somewhat complicated.

One solution is first merging small regions into one entire scene prediction and then evaluating the entire scene \cite{Wang2017}.
However, the final result is largely affected by the merging algorithm, and it is difficult to evaluate the pure instance segmentation performance.

Another way is evaluating the instance segmentation in small regions \cite{sung2018}; however, this is not desirable owing to the following reason.
As shown in Figure \ref{fig:size_of_grid}, small regions often contain only one instance for a certain category, and in such cases, the resulting semantic segmentation is sufficient for instance segmentation because it is not necessary to distinguish objects belonging to the same class.
When the input regions are too small and there is only one object in each region, it is unnecessary to distinguish the object, and thus, instance segmentation does not have to be conducted.
Conversely, when there are many objects belonging to the same class, it is necessary to consume larger regions in order to evaluate an instance segmentation.
Consuming large regions is also challenging because it is necessary to consume a large number of points to avoid a sparse input, which decreases the performance of certain models including PointNet \cite{Qi2016,Qi2017b}. Handling dense point clouds is also helpful in the application of instance segmentation.
However, as an example, the Similarity Group Proposal Network (SGPN) \cite{Wang2017} calculates the similarities for each pair of points and its space complexity is $\mathcal{O}(N_p^2)$ for $N_p$ number of points, which makes it difficult to consume large point clouds.
Thus, a memory efficient method is required.

Moreover, there are certain problems in existing metrics.
The method in \cite{Wang2017} was evaluated using the mean average precision (mAP), which has a characteristic in that the effect of false positives with low confidence scores is small. 
Although this property is appropriate for tasks such as object detection where multiple candidates with overlaps are allowed, or retrieval where the rank of the output is important, this property is not suitable for instance segmentation.
Because outputs are objects without an overlap for each point, we need to equally evaluate the outputs of each point regardless of the confidence score.
In addition, when evaluating instance segmentation, we focus on whether two objects are properly distinguished, and whether one object is incorrectly split.
However, these failures cannot be distinguished from a misclassification when performing evaluations using existing metrics, and a misclassification is often the main factor of decreasing mAP.


In this study, we first experimentally show our claim that evaluating instance segmentation in small regions with existing metrics is inappropriate and reveal the problem using metrics that has not been investigated in previous studies.
Then, we propose a novel instance segmentation method with small space complexity that enables the consumption of large regions.
Our loss function learns a one-to-one mapping from an input feature space to an embedding space, where embeddings from the same instance form a cluster, and we can distinguish instances by clustering at the test time.
Because our method does not have to handle point pairs, the space complexity is $\mathcal{O}(N)$ and is scalable to the number of points.
We show that the proposed memory efficient method outperforms other state-of-the-art methods.

In addition, we demonstrate that consuming small regions and evaluating them by using existing metrics is not appropriate.
This fact has been overlooked by previous research, so we propose a novel metric that can evaluate it correctly for the first time.
Our metric is based on inclusion, which is the relationship of one set being a subset of another. Using the proposed metric, we can evaluate the pure performance regardless of the size of the regions, categories, or confidence scores.
We can also analyze the types of errors quantitatively.

We conducted extensive experiments to reveal the effect of the size of the regions and the density of the points on the instance segmentation performance and showed that consuming a large number of points increases the performance for large regions.

The key contributions of this study are as follows:

\begin{itemize}
    \item We propose a new loss function that learns to push embeddings for each instance to be clustered and is scalable to the number of points; we also experimentally demonstrated that the proposed method outperforms existing methods.
    \item We reveal the problems associated with existing metrics, including the fact that they are affected by the size of inputs or categories, which have been overlooked in previous research.
    \item We propose a novel metric that is not affected by these factors and can evaluate instance segmentation performance correctly.
\end{itemize}

\section{Related Work}
\label{sec:related_works}

\subsection{Feature Extraction on 3D Point Clouds}
\label{sec:rel_point_cloud}
The effective handling of point clouds is challenging because they are unordered, non-uniformly distributed data.

Methods to extract features from point clouds can be roughly classified into two approaches, namely, describing local features \cite{rusu2008aligning,rusu2009fast} and describing relationships among multiple points \cite{drost2010model,choi2012voting}.

PointNet \cite{Qi2016}, which addresses the problem of unordered data by using symmetric functions, and PointNet++ \cite{Qi2017b}, which stacks PointNets and is able to handle local features, have made recent breakthroughs in deep learning on point clouds.
We use PointNet and PointNet++ as feature extractors in this study.

\subsection{Instance Segmentation}
\label{sec:rel_segmentation}
Segmentation is a task of labeling each minimum element in the data such as a pixel or point.
In particular, labeling the category of each element and distinguishing objects belonging to the same category are called instance segmentation against semantic segmentation.

\paragraph{2D Images}
Many studies on instance segmentation on images have been recently published \cite{dai2016instance,li2016fully,He2017,DeBrabandere2017,Kong2017,Novotny2018},
and Novotny \etal. \cite{Novotny2018} classified instance segmentation into two approaches, \textit{propose \& verify} (P\&V) and \textit{instance coloring} (IC).
P\&V is an approach that first proposes candidates of objects based on their objectness and then verifies whether an object is a candidate. This is currently a popular approach in the field of object detection \cite{Ren2015a} and instance segmentation \cite{He2017} on images.
Although P\&V approaches have achieved significant success in image segmentation, they have weaknesses in that object candidates are approximations of the object shapes, and a second-stage to refine the candidates is necessary for segmentation, such as Mask R-CNN \cite{He2017}, and thus, the network architecture tends to be complex.

Approaches labeling an object identifier directly to each pixel are called IC, and some studies have been conducted in this area for image segmentation \cite{DeBrabandere2017,Kong2017,Novotny2018}.
Brabandere \etal. \cite{DeBrabandere2017} proposed a discriminative loss function that learns a mapping to an embedding space where the embeddings form clusters for each object.
The loss function is simple and efficient but has some shortcomings, as described in Section \ref{sec:prior_dic_loss}.

We choose an IC-based approach because the architecture tends to be simpler, and it is thus expected to be computationally efficient.

\paragraph{3D Point Clouds}
SGPN \cite{Wang2017} and deep functional dictionaries (DFD) \cite{sung2018} have tackled instance segmentation on 3D point clouds.
SGPN first predicts similarities for every pair of points that describes whether two points belong to the same object and then merges points to instance proposals by considering a pair of points with a similarity higher than a certain threshold as being contained in the same object.
Although it is a pioneering work of instance segmentation on points clouds, the space complexity of the similarity matrix is proportional to the square of the number of points and cannot handle too many points. We discuss this problem in Section \ref{sec:analyze}.
Thus, input scenes are split into \SI{1}{\metre} square regions, and the results are then aggregated for each region using a heuristic algorithm. However,
the final performance depends on the merging algorithm, as described in Section \ref{sec:introduction}, and applying the method for every small region is computationally inefficient.

Recently, Sung \etal. \cite{sung2018} proposed a general method called DFD that produces a dictionary of the probe functions.
The authors proposed a general framework that learns a mapping from the shape to the dictionary.
Each atom of the dictionary can be associated with semantics, instances, or something else based on the task and constraint.
A performance comparable to that of state-of-the-art techniques was achieved on S3DIS, but the authors evaluated its performance for each small region. Thus, this evaluation has certain problems, as discussed in Section \ref{sec:introduction}.

\subsection{Embedding Learning}
\label{sec:prior_dic_loss}
Our method performs instance segmentation by first learning the feature embeddings for each point such that the diameter of the embedding cluster corresponding to the same object is small compared to the distance among clusters from different objects; then, clustering is conducted in the embedding space.
Such a feature learning method that trains the embedding to minimize the distance between embeddings with the same semantics while maximizing the distance between embeddings with different semantics is widely used in category classification \cite{chopra2005learning,weinberger2009distance} and similarity learning \cite{koestinger2012large,schroff2015facenet}.
This concept has been used for recent instance segmentation studies on images such as those on discriminative loss \cite{DeBrabandere2017}, \cite{Kong2017}. Inspired by this, we propose a novel instance segmentation method that overcomes the discriminative loss problem. 

Discriminative loss $L$ consists of $L_{\mathrm{var}}$, which makes the distance between points and centroids of the corresponding cluster smaller than $\delta_v$; $L_{\mathrm{dist}}$, which makes the distance between cluster centroids larger than $\delta_d$; and a regularizer $L_{\mathrm{reg}}$, which prevents the feature norms from diverging. Here, $L$ is written as follows:
\begin{align}
    L_{\mathrm{var}} &= \frac{1}{C}\sum_{c=1}^C\frac{1}{N_c}\sum_{i=1}^{N_c}[\|\bm{\mu_c}-\bm{x_i}\|-\delta_v]_+^2 \\
    L_{\mathrm{dist}} &= \frac{1}{C(C-1)}\sum_{c_A\neq c_B}[2\delta_d-\|\bm{\mu_{c_A}} - \bm{\mu_{c_B}}\|]_+^2 \label{eq:disc_recall_dist} \\
    L_{\mathrm{reg}} &= \frac{1}{C}\sum_{c=1}^C\|\bm{\mu_c}\|  \label{eq:disc_recall_reg} \\
    L &= L_{\mathrm{sem}} + \alpha L_{\mathrm{var}} + \beta L_{\mathrm{dist}} + \gamma L_{\mathrm{reg}}, \label{eq:total_loss}
\end{align}
where $C$ denotes the number of clusters, and $\bm{\mu_c}$ and $N_c$ are the centroid and number of points of cluster $c$, respectively, $\bm{x_i}$ is the embedding, $L_{\mathrm{sem}}$ is the softmax cross entropy loss of the category classification, $\|\cdot\|$is the Euclidean norm in the feature space, and $[x]_+ = \max(0, x)$.
When we conduct instance segmentation, we apply clustering on the learned embedding space.
When we set $\delta_d \ge \delta_v$ and the learned embedding space satisfies $L_{var} = L_{dist} = 0$, we can guarantee that all points whose distances from a point are smaller than $\delta_v$ belong to the same object.

However, there are some drawbacks in this original formulation of discriminative loss.
First, it is difficult to select the hyperparameter $\beta, \gamma$ that balances the weights of $L_{\mathrm{reg}}$ and $L_{\mathrm{dist}}$. The optimization is
hyperparameter-sensitive because $L_{\mathrm{reg}}$ attempts to reduce the distances between points (i.e., make them closer), whereas $L_{\mathrm{dist}}$ attempts to increase the distance between points (i.e., make them more distant).
Empirically, it turns out that $\gamma$ should be about 100-times smaller than $\alpha, \beta$, and seeking such balance is an cumbersome task.
Moreover, when we concatenate the learned feature to other features such as the raw coordinates of a point, we need to arrange the scale of the features such that both features are effective for clustering. However, it is difficult to arrange the scale because the norms of the feature are different among feature spaces. In contrast, when we normalize each feature after we learn the feature space, we cannot distinguish between points with the same unit vector and a different norm.

In the following sections, we propose a novel embedding method that solves these problems.

\section{Method}
\label{sec:methods}

\subsection{Proposed Feature Embedding}
\label{sec:metric_method}
In this section, we describe the proposed feature learning method.
As described in a previous section, Equation (\ref{eq:disc_recall_dist}) in $L$ attempts to increase the distances between different clusters while minimizing the norms of the feature using Equation (\ref{eq:disc_recall_reg}). Thus, $L$ is sensitive to the hyperparameters $\beta$ and $\gamma$ that balance these conflicting losses. Moreover, it is difficult to combine a learned feature with other features for clustering.

\begin{figure}
    \begin{center}
        \includegraphics[width=5.5cm]{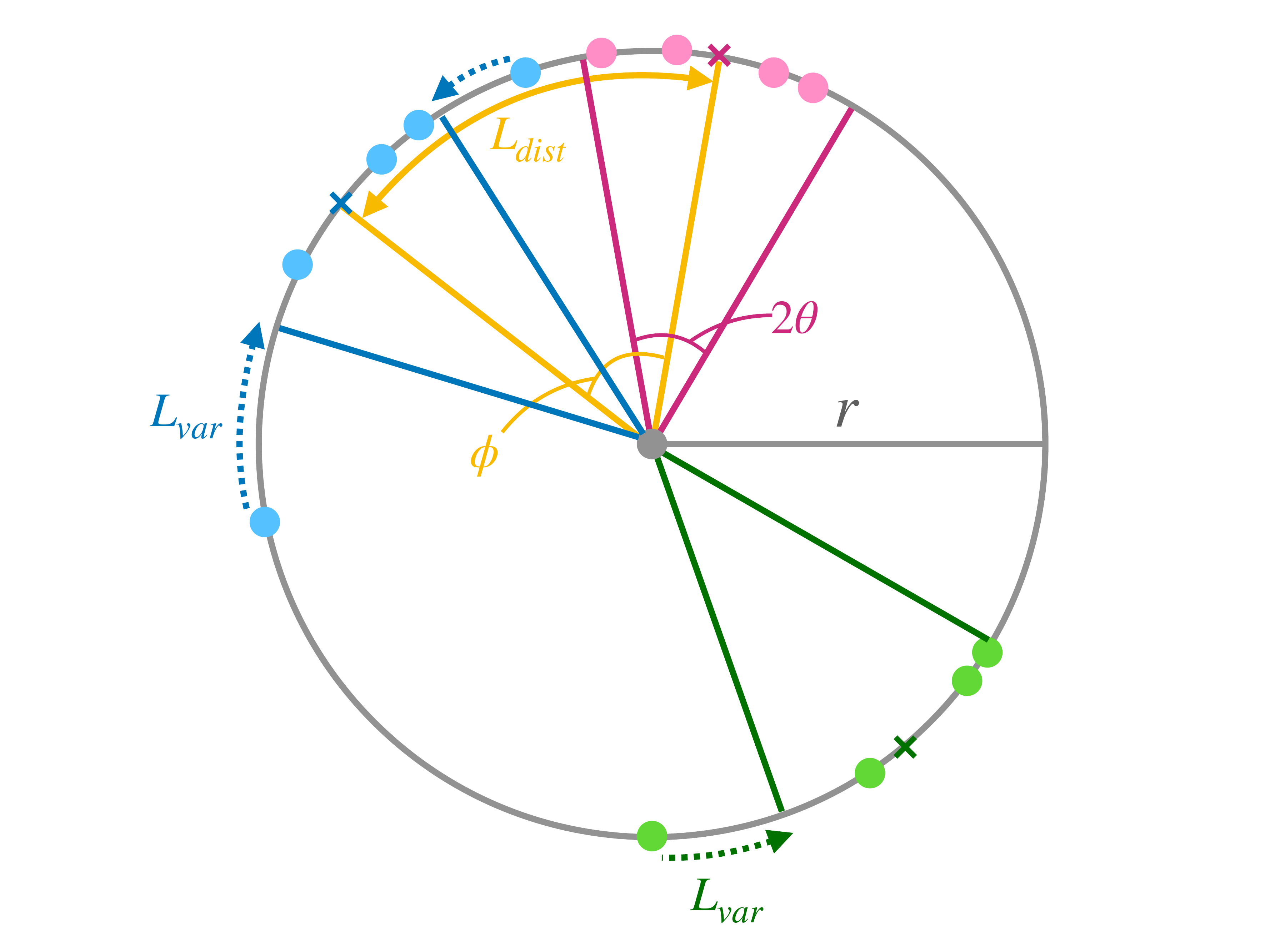}
    \end{center}
    \caption{\label{fig:cosine}Overview of the proposed feature learning method. Each point ($\bullet$) represents one feature embedding, and the points with the same color belong to the same object. The crosses ($\times$) are cluster centroids. As the training progresses, points with the same colors move to a nearby spot and clusters move away from each other.}
\end{figure}

In this study, we overcome these difficulties by restricting the features to a unit hypersphere and the learning of the feature space based on a cosine similarity instead of the Euclidean loss.
We present an overview of our method in Figure \ref{fig:cosine}, where each point ($\bullet$) represents one feature embedding, points with the same color belong to the same object, and the cross ($\times$) indicates the cluster centroid. Moreover, $\theta$ and $\phi$ satisfy  $\delta_v=\cos(\theta)$ and $\delta_d < \cos(\phi)$, respectively.

Using the cosine similarity between two embeddings $\bm{x}_i, \bm{x}_j$, which is calculated as $s(\bm{x}_i, \bm{x}_j) = \frac{\bm{x}_i^T\bm{x}_j}{\|\bm{x}_i\|\|\bm{x}_j\|}$, the proposed loss function is written as follows:
\begin{align} \label{eq:cosine}
    L_{\mathrm{var}} &= \frac{1}{C}\sum_{c=1}^C\frac{1}{N_c}\sum_{i=1}^{N_c}[\delta_v - s(\bm{\mu}_c, \bm{x}_i)]_+ \\
    L_{\mathrm{dist}} &= \frac{1}{C(C-1)}\sum_{c_A\neq c_B}[s(\bm{\mu}_{c_A}, \bm{\mu}_{c_B}) - \delta_d]_+ \\
    L &= L_{\mathrm{sem}} + \alpha L_{\mathrm{var}} + \beta L_{\mathrm{dist}},
\end{align}
where $\delta_d$ and $\delta_v$ satisfy $\delta_d \ll \delta_v \approx 1$ so that $s(\bm{\mu}_c, \bm{x}_i)$ becomes larger than $s(\bm{\mu}_{c_A}, \bm{\mu}_{c_B})$.
In addition, we use the absolute error of the $[\cdot]_+$ terms instead of the squared error adopted in \cite{DeBrabandere2017} because the norm of the $[\cdot]_+$ terms is smaller than 1 and the squared errors become considerably smaller when these terms are near zero.

When the angle between an embedding and its cluster centroid is larger than $\theta$, $L_{\mathrm{var}}$ attempts to reduce the distance between the embedding and the centroid. In addition, $L_{\mathrm{dist}}$ attempts to increase distance between cluster centroids when the cosine similarity is larger than $\delta_d$. In an image recognition study (\cite{Kong2017}), the feature was also learned using the cosine similarity using a unit hypersphere. However, in that study, similarities between all pairs of points were calculated, whereas
our method only considers the similarities between points and the corresponding cluster centroids. Thus, our method is considerably more computationally effective. 

Compared to \cite{DeBrabandere2017}, the advantages of our method are as follows:
\begin{itemize}
\item We do not need to consider the scale of the feature space and thus, we can omit $L_{\mathrm{reg}}$ and do not need to consider the balance between $\beta$ and $\gamma$.
\item Because the embeddings are guaranteed to have a unit norm, it is easy to combine the learned embeddings to other features.
\end{itemize}

We learn the mapping from the feature space to the embedding space by adding one fully connected layer.

\subsection{Computational Complexity}
\label{sec:analyze}
In Section \ref{sec:rel_segmentation}, we discussed the fact that one of the problems of the existing IC-based instance segmentation method SGPN \cite{Wang2017} is that it requires a large space complexity. 
Because our method and SGPN require only a few extra layers in the feature extractor, and thus, the number of iterations for training is nearly the same, we focus on analyzing the detailed computation complexity of the loss functions of SGPN and our method.
In the following section, we denote the batch size as $B$, the number of points as $N_p$, the number of points in a cluster $c$ as $N_c$, and the dimensions of the input feature space and embedding space as $d_f$ and $d_e$, respectively. We also write the input and embedded features of the $i$-th point as $\bm{f_i}\in \mathbb{R}^{d_f}$ and $\bm{h_i}\in \mathbb{R}^{d_e}$, respectively.

In SGPN, the similarity $S_{ij}$ between the $i$-th and $j$-th points is calculated as
\begin{equation}
  S_{ij} = \|\bm{f_i}-\bm{f_j}\| = \sqrt{\|\bm{f_i}\|^2-2\bm{f_i}\bm{f_j}+\|\bm{f_j}\|^2}.
\end{equation}
Because the method calculates $S_{ij}$ for all pairs of points, the space complexity of the similarity matrix is $\mathcal{O}(BN_p^2d_f)$.
As for the time complexity, because we need to evaluate $\|f_i\|^2$ for each $i$ and $f_if_j$ for each pair $(i, j)$ to calculate $S_{ij}$, the time complexity is $\mathcal{O}(BN_p^2d_f)$.

In contrast, the proposed loss function obtains $d_e$-dimensional embedded features and calculates the cosine similarity between each point and its cluster centroid, and between each pair of cluster centroids. Therefore, the space complexity for the embeddings of each point is $\mathcal{O}(BN_pd_e)$, and the computation complexity is $\mathcal{O}(B(N_p+C^2)d_e)$; however, this order is equivalent to $\mathcal{O}(BN_pd_e)$ because $C \ll N_p$ in most cases. Both complexities are linear in $N_p$.
Because we use $N_p = 2^n (n=12, 13, 14), d_e=2^5$ in the experiment, the proposed method can calculate the loss function with a smaller space/time complexity than SGPN.

\subsection{Clustering}
We describe our feature learning method in Section \ref{sec:metric_method}.
In this section, we explain the clustering method applied to the learned feature space to conduct instance segmentation.

The requirements for the clustering method are as follows:

\begin{itemize}
    \item The number of clusters is variable.
    \item The clustering result is robust to outliers.
    \item The clustering does not fail even when the number of points in each cluster has a large variety.
\end{itemize}

In this study, we adopt the density-based spatial clustering of applications with noise (DBSCAN) \cite{ester1996density}, which satisfies these requirements.
DBSCAN is a density-based clustering method that first calculates the densities of points based on the number of neighboring points and then constructs clusters by considering a continuous region with a density of above a certain threshold as a single cluster.
The number of clusters of the output of DBSCAN can vary, and DBSCAN is robust to outliers because it accepts the noise points that do not belong to any clusters.

We apply this clustering to the embeddings, which are predicted as the same category.
We concatenate the learned embeddings using the normalized coordinates of the point as the input for the clustering method.
In addition, some clusters consist of a very small number of points. Because such clusters are false positive in most cases, we handle such clusters as points in that they do not belong to any cluster in the evaluation.

\section{Evaluation Metrics}
\label{sec:metric}
As described in Section \ref{sec:introduction}, existing metrics of instance segmentation are affected by the misclassification, the confidence of prediction, and the size of the regions.
Therefore, we propose a novel evaluation metric that focuses on the distinct ability of the objects regardless of the confidence or semantics, and which can be used for any sized input region.

When we neglect semantic errors, we can observe four patterns for each prediction output:

\begin{itemize}
    \item There is a corresponding ground truth (GT) for the prediction output (true positive (TP)).
    \item The prediction output covers some part of a GT (partial detection (PD)).
    \item The prediction output contains more than one GT (false merging (FM)).
    \item There is no corresponding GT (false positive (FP)).
\end{itemize}

\begin{figure}
    \begin{center}
        \includegraphics[width=7cm]{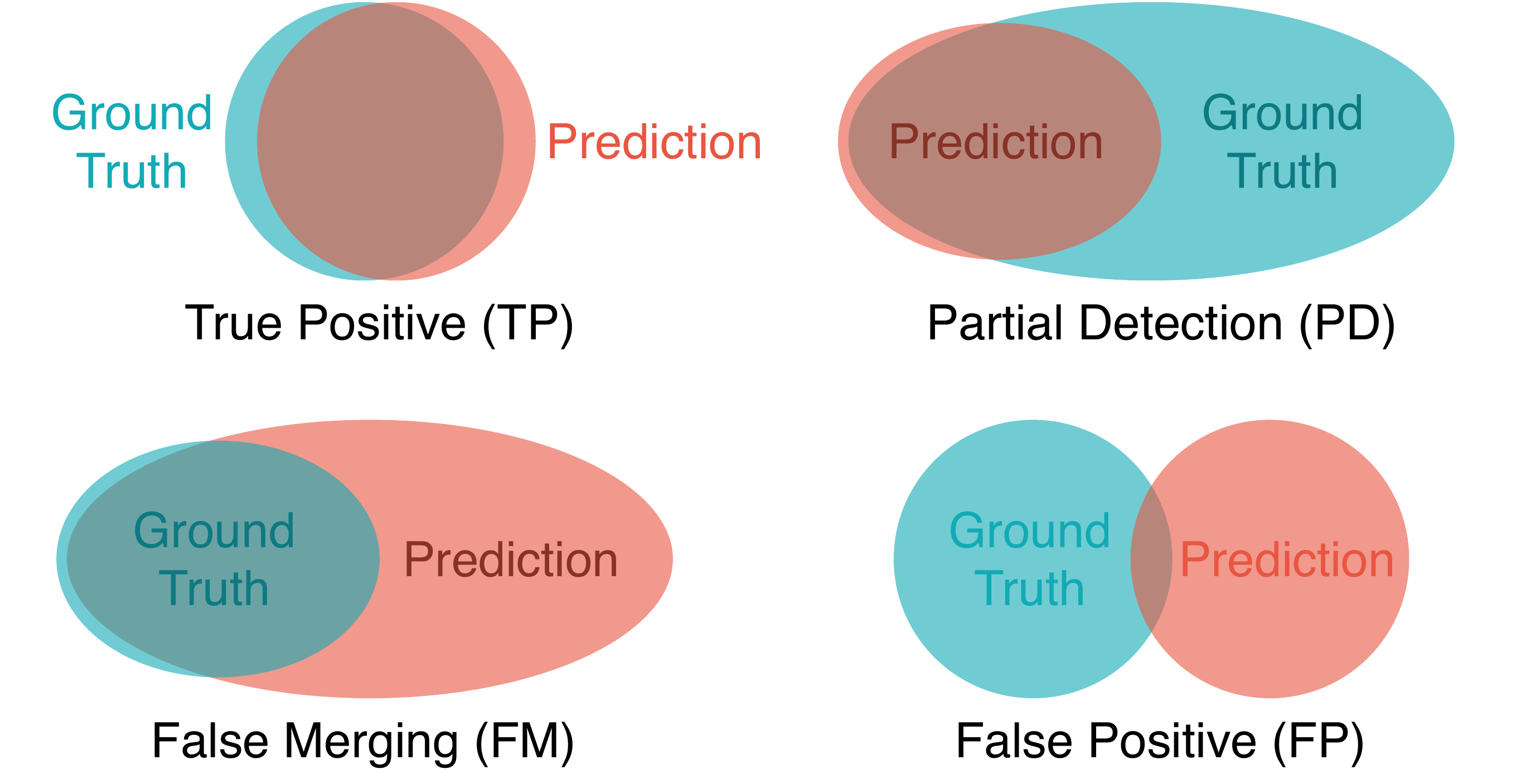}
    \end{center}
    \caption{\label{fig:errors} Error patterns for instance segmentation}
\end{figure}

Figure \ref{fig:errors} shows a diagram of these four error patterns.

Note that one prediction output can fulfill more than one of the patterns even though each point corresponds to exactly one GT and one prediction.
For example, one prediction output, 90\% of which is contained in a GT, can cover other small GTs with the remaining 10\%.
In particular, PD and FM are characteristics of instance segmentation.

To formulate these patterns, we define "intersection over a set (IoS)", which describes the part of an object $A$ that is contained in an object $B$ as follows:
\begin{align}
    \label{eq:ios}
    IoS(A, B) = \frac{N(A\cap B)}{N(A)}
\end{align}
where $N(X)$ denotes the number of points in $X$, and object $A$ is considered to be contained in object $B$ when $IoS(A, B)$ exceeds a certain threshold $t$.
Note that IoS() is an asymmetric function and $A$ cannot be contained in more than one object when we set $t>0.5$.

Using this IoS, we can establish the proposed metrics as follows.
We first calculate a map from the GTs to the prediction outputs ($gt2pred$) that describes which prediction outputs are contained in each GT, and conversely calculate a map from the prediction outputs to the GTs ($pred2gt$) that describes which GTs are contained in each prediction. Note that $gt2pred$ and $pred2gt$ are not exclusive.
Then, we label each prediction for at least one of the patterns.
As for $gt2pred$, for each GT $g$, we can obtain a list of prediction outputs corresponding to $g$ ($g2p$).
If a prediction $p$ is on the list that also contains $g$ itself, $p$ is considered as TP; otherwise, $p$ is considered as PD because $p$ is a subset of $g$.

In contrast, for $pred2gt$, with each prediction output $p$, we can obtain a list of GTs corresponding to $p$ ($p2g$).
If a GT $g$ on the list does not contain any data and $p$ is not labeled as PD in the last process, $p$ is considered as FP.
Otherwise, if $g$ also contains $p$ itself, it must be labeled as TP in the last process owing to its symmetry.
Here, $g$, which does not contain $p$, is considered as FM because $g$ is a subset of $g$ in this case.
We define the ratio of TP to the number of predictions as \textit{precision} and the ratio of TP to the number of GTs as \textit{recall}; we define the F-score as their harmonic mean.
We also evaluate the error patterns based on the ratio of PD, FM, and FP to the number of predictions.
Because we ignore the semantic segmentation in the calculation, one prediction output can be TP even if its predicted semantics are incorrect.
This proposed metric does not depend on the semantics, confidence, or size of the input regions. Therefore, we can evaluate the pure performance of the instance segmentation.
The procedure explained above can be written as Algorithm \ref{Algorithm:proposed_criteria}.

\algnewcommand\algorithmicswitch{\textbf{switch}}
\algnewcommand\algorithmiccase{\textbf{case}}

\algdef{SE}[SWITCH]{Switch}{EndSwitch}[1]{\algorithmicswitch\ #1\ \algorithmicdo}{\algorithmicend\ \algorithmicswitch}%
\algdef{SE}[CASE]{Case}{EndCase}[1]{\algorithmiccase\ #1}{\algorithmicend\ \algorithmiccase}%
\algtext*{EndSwitch}%
\algtext*{EndCase}%

\alglanguage{pseudocode}
\begin{algorithm}[h]
\small
\caption{Criteria for instance segmentation}
\label{Algorithm:proposed_criteria}
\begin{algorithmic}[1]

\Procedure{$\mathbf{AggregateResults}$}{$GT$, prediction $P$}
    \State arr $gt2pred$[len($G$)][] \Comment{\emph{map from GT to preds}}
    \State arr $pred2gt$[len($P$)][] \Comment{\emph{map from pred to GTs}}
    \For {each $g$ in $G$}
        \For {each $p$ in $P$}
            \If {$s(g\cap p)/s(g) > t$}
                \Comment{\emph{$g$ is included in $p$}}
                \State $pred2gt$.append($g$)
            \EndIf
            \If {$s(g\cap p)/s(p) > t$}
                \Comment{\emph{$p$ is included in $g$}}
                \State $gt2pred$.append($p$)
            \EndIf
        \EndFor
    \EndFor
    \State \textbf{Summarize($gt2pred$, $pred2gt$)}
\EndProcedure

\Procedure{$\mathbf{Summarize}$}{$gt2pred$, $pred2gt$}
    \State $results$[len($P$)][] \Comment{\emph{2D array to store predictions}}
    \For {each $g2p$ in $gt2pred$}
        \For {each $p$ in $g2p$}
            \If {$g$ in $pred2gt[p]$}
                \State $results[p]$.append("TP") \Comment{\emph{true positive}}
            \Else
                \State $results[p]$.append("PD") \Comment{\emph{partial detection}}
            \EndIf
        \EndFor
    \EndFor
    \For {each $p2g$ in $pred2gt$}
        \If {$\mathrm{len}(p2g)==0$ and $results[p] == []$}
                \State $results[p]$.append("FP") \Comment{\emph{false positive}}
        \EndIf
        \For {each $g$ in $p2g$}
            \If {$p$ not in $gt2pred[g]$}
                \State $results[p]$.append("FM") \Comment{\emph{false merging}}
            \EndIf
        \EndFor
    \EndFor
\EndProcedure
\Statex
\end{algorithmic}
  \vspace{-0.4cm}
\end{algorithm}

\section{Experiments}
\label{sec:experiments}
In this section, we conduct experiments to compare our method with existing methods in order to demonstrate its effectiveness and to show that existing evaluation metrics of instance segmentation in small regions are inappropriate.
We then clearly distinguish errors of misclassification and splitting instances by using our proposed evaluation metric, which cannot be achieved using existing evaluation metrics such as mAP.
Moreover, we evaluate the relationships between the size of the split regions and the instance segmentation performance to validate our assumption that evaluating instance segmentation methods using existing evaluation metrics for small regions is inappropriate. 

\subsection{Datasets and Setup}
\label{sec:experiment_settings}
We use the Stanford large-scale 3D Indoor Spaces Dataset (S3DIS) \cite{Armeni2016}. S3DIS consists of 270 indoor scenes scanned from six areas and 13 objects. We use 203 scenes for training and the remaining 67 scenes for evaluation.

Although PointNet++ \cite{Qi2017b} and SGPN \cite{Wang2017} have been used to evaluate methods by splitting the input scene horizontally into small regions, such as \SI{1}{\meter} square regions, we conducted additional experiments using larger regions as input. This is because one of our aims is to construct a method that can be applied to wide regions with a greater number of points. During each training iteration, we randomly sample subregions with a fixed size from each scene, and then randomly sample a fixed number of points from the sampled subregion as the input. In the following experiments, the region size is \SI{1}{\meter} square, and the number of points is 4,096 unless otherwise noted.
Each point has a nine-dimensional normalized feature consisting of RGB values, relative coordinates in the subregion, and absolute coordinates in the room.
For data augmentation, we apply random noise to some of the input features.

The number of objects in the dataset differs significantly among categories. For example, the number of objects of the category with the largest number of objects is 55 times as large as the number for the category with the smallest number of objects. To eliminate the effect of this imbalance, we weight the miscategorization cross-entropy loss as the weight corresponding to the category with a small number of points weighted as a large value.

Although we can apply our embedding learning method to any feature extractors, we use PointNet (PN) \cite{Qi2016} and PointNet++ (PN++) \cite{Qi2017b} as feature extractors for our experiments.
We use 131-dimensional features consisting of 128-dimensional features extracted using the feature extractor and three-dimensional RGB features as the input for feature embedding.
The 131-dimensional features are then passed through a fully connected layer, which produces 32-dimensional embeddings.

We use the Adam \cite{kingma2014adam} optimizer with an initial learning rate of 0.001 and a batch size of 32. We train our network for 6,000 steps, and the learning rate is divided by 10 at the 4,500th step.
We set the hyperparameters for our method as $\delta_v = 0.9, \delta_d = 0.4, \alpha = \beta = 0.5$.

\subsection{Evaluation on Existing Evaluation Metrics}
\label{sec:metric_result}
In this section, we evaluate the proposed embedding learning method using existing instance segmentation metrics.
We compare our method with SGPN \cite{Wang2017} and DFD \cite{sung2018}, which have been found to exhibit the highest accuracy for this task.
The scores for these two methods are reported in \cite{sung2018}.

Following \cite{sung2018}, we chose the proposal recall \cite{proposal_recall} as the evaluation metric and used PointNet as a feature extractor for a fair comparison.
The proposal recall is calculated as follows: first, for each GT object, we select the predicted object with the highest intersection over union (IoU) regardless of the category of the object and consider the output as a true positive when the IoU is higher than a certain threshold (we chose a value of 0.5). The ratio of the number of true positives is then calculated with respect to the number of GTs.
Because the number of objects for each category is unbalanced, we evaluated both the mean of the proposal recall of 12 categories, except the 'clutter' class (mean), and the overall proposal recall regardless of the categories (total).
Note that the overall proposal recall (total) can be high even if the model overfits some of the categories with many instances and ignores the categories with fewer instances, and thus, it may not be reliable.
However, DFD, which does not use category information for training, cannot solve the imbalance problem between categories, and thus, it was necessary to add the total proposal recall.

Moreover, to validate our argument that instance segmentation on small regions is a substantially semantic segmentation because there is often only one instance in the region, 
we also evaluated the result obtained using semantic segmentation model (SemSeg), which never splits objects belonging to the same category.
We also report the score obtained when using PointNet++ (PN++) instead of PN as the feature extractor; however, we do not compare PN++ with PN as it would not make for a fair comparison.

Table \ref{table:compare_result} shows a comparison of our methods with existing methods as well as the obtained semantic segmentation results.
We can see that our method with PointNet (PN) outperforms existing methods in terms of the mean proposal recall by a large margin, and the use of PointNet++ leads to a considerably better score.
As described earlier, DFD achieves a high total score; however, its mean score is low, which means that the model ignores categories with fewer instances.
In addition, for some categories such as a ceiling, floor, and beam, a mere semantic segmentation result achieves a very high score because there is essentially only one instance of such categories. This result supports our argument that semantic segmentation results affect the instance segmentation performance, and an evaluation in small regions using existing metrics is inappropriate.

\begin{table*}
    \caption{\label{table:compare_result} Comparison with existing methods (\cite{Wang2017,sung2018}) by proposal recall [\%]}
    \begin{center}
        \small
        \begin{tabular}{c|cccccccccccc|c|c}
            \hline
            \multirow{2}{*}{method} & ceil- & \multirow{2}{*}{floor} & \multirow{2}{*}{wall} & \multirow{2}{*}{beam} & col- & win- & \multirow{2}{*}{door} & \multirow{2}{*}{table} & \multirow{2}{*}{chair} & \multirow{2}{*}{sofa} & book- & \multirow{2}{*}{board} & \multirow{2}{*}{mean} & \multirow{2}{*}{total} \\
            & ing &  &  &  & umn & dow & & & & & case & & & \\\hline
            SGPN \cite{Wang2017} & 67.0 & 71.4 & 66.8 & 54.5 & 45.4 & 51.2 & 69.9 & 63.1 & \textbf{67.6} & 64.0 & 54.4 & \textbf{60.5} & 61.3 & 64.7 \\
            DFD \cite{sung2018} & 95.4 & \textbf{99.2} & \textbf{77.3} & 48.0 & 39.2 & 68.2 & 49.2 & 56.0 & 53.2 & 35.3 & 31.6 & 42.2 & 57.9 & \textbf{69.1} \\
            SemSeg & 95.8 & 95.2 & 61.7 & \textbf{89.3} & 50.0 & 76.6 & 65.7 & 60.2 & 44.1 & 16.6 & 40.6 & 45.6 & 61.8 & 59.4 \\
            Ours (PN) & \textbf{95.9} & 94.6 & 64.5 & \textbf{89.3} & \textbf{61.3} & \textbf{83.3} & \textbf{75.0} & \textbf{64.0} & 55.0 & \textbf{70.8} & \textbf{55.6} & 50.8 & \textbf{71.7} & 68.8 \\\hline
            Ours (PN++) & 96.2 & 94.1 & 65.6 & 87.8 & 71.4 & 81.0 & 82.6 & 68.8 & 60.9 & 68.4 & 63.2 & 67.2 & 75.6 & 72.7 \\\hline
        \end{tabular}
    \end{center}
\end{table*}

\subsection{Evaluation of the Proposed Evaluation Metrics}
We then evaluated our method with DFD, which outperforms SGPN, using the proposed evaluation metrics.
Some predicted objects consist of a very small number of points.
Because such predicted objects are often false positives, we set a threshold and use the predicted objects with a number of points larger than the threshold as the targets for evaluation.
There is a trade-off between \textit{precision} and \textit{recall}, which we introduced in Section \ref{sec:metric}. As the threshold decreases, \textit{precision} decreases while \textit{recall} increases.
We fix $t=0.75$ for the IoS defined by Equation \ref{eq:ios} and search for the threshold that can obtain the highest F-score.
As a result, we use a threshold of 150 for the DFD and 35 for our method. The DFD shows a larger threshold, which implies that it outputs noisy small predicted objects that are false positives.

Furthermore, as discussed in Section \ref{sec:metric_result}, the output of the semantic segmentation model (SemSeg), which never splits objects belonging to the same category, achieves high scores using the existing metrics for instance segmentation. This occurs in some categories in which multiple objects seldom exist in a single subregion.
We evaluated the results of semantic segmentation using our evaluation metrics to demonstrate whether this problem was solved.

\begin{figure}
    \begin{center}
        \includegraphics[width=8.2cm]{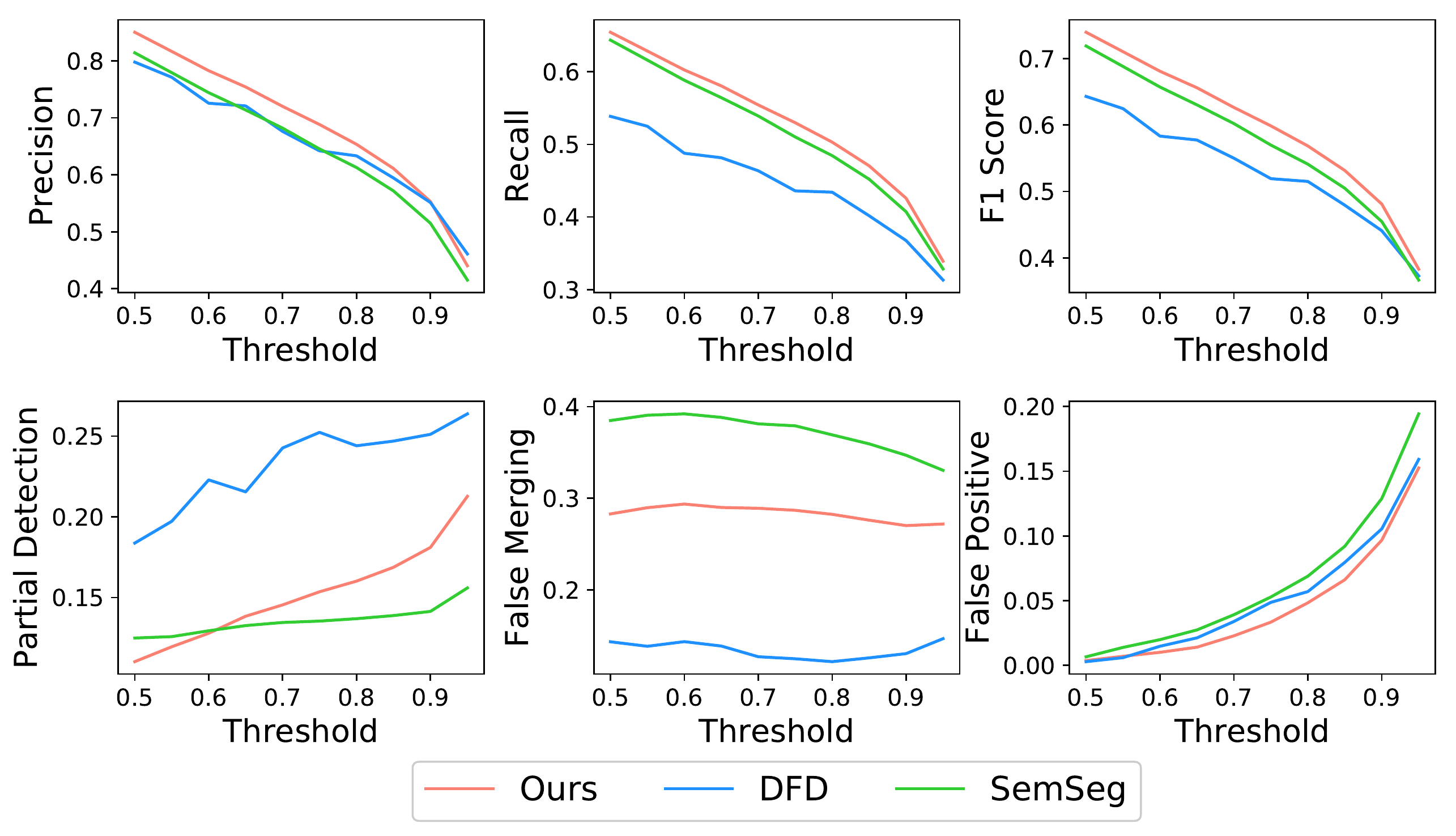}
    \end{center}
    \caption{\label{fig:criteria} Comparison of our method and \cite{sung2018} using the proposed evaluation metrics}
\end{figure}

We plot the results when varying $t$ of IoS from 0.5 to 0.95 in Figure \ref{fig:criteria}. Note that precision, recall, and f1-score evaluate performance, whereas partial detection, false merging, and false positive represent types of mistakes. We can see that, although the semantic segmentation model shows a high score for the existing metrics, specifically for \textit{recall}, the false merging score is quite high. This is because semantic segmentation outputs one prediction per category at most; this is why the partial detection of the semantic segmentation model is low.
To the best of our knowledge, this fact is revealed by the proposed metrics for the first time.
We can also observe fine patterns and the property of instance segmentation errors when eliminating semantic errors, which cannot be obtained using existing evaluation metrics.
For example, for DFD, most errors arise from partial detection whereas false merging is the dominant cause in our method.
Such information is useful not only for analyzing and improving the model but also for applying an ensemble of models when considering the characteristics of each model.

\subsection{ Effect of Region Size and Number of Points}
\label{sec:metric_result_grid_size}
In this section, we analyze the effects of the input region size and the number of points.
We varied the number of points from 4,096 to 16,384 and the size of the regions from \SI{1}{\metre} square from \SI{4}{\metre} square. 
Note that the instance segmentation results can be also affected by the density of the points. The settings with 1,024 points and \SI{1}{\metre} square, 4,096 points and \SI{2}{\metre} square, and 16,384 points with \SI{4}{\metre} square have the same density.

\begin{figure}
    \begin{center}
        \includegraphics[width=8cm]{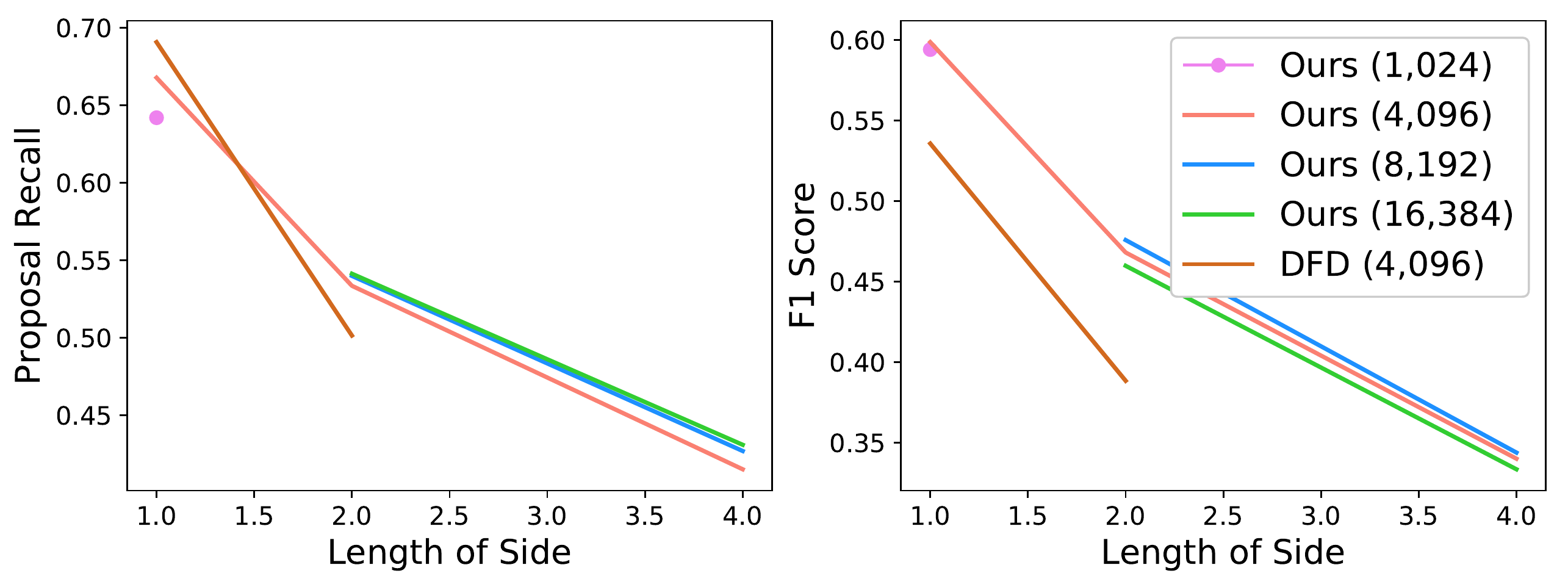}
    \end{center}
    \caption{\label{fig:result_grid_size} Effect of region size and number of points}
\end{figure}

Figure \ref{fig:result_grid_size} shows the proposal recall and the F-score values for each setting.
Even if the density is the same, the score of the instance segmentation decreases with the size of the region.
As discussed in Section \ref{sec:introduction}, when the input region is small, we do not need to distinguish the objects because the number of different objects of the same category is small. Therefore, the task becomes difficult as the input region increases and the apparent score decreases.
In particular, the F1 score of DFD is significantly decreased compared with the proposal recall; one reason for this is that the proposal recall does not penalize false positives and cannot reveal the weakness of the DFD model that the instance segmentation results are quite noisy.
Moreover, the figure shows that the density of points does not considerably affect the instance segmentation performance.

Although we can apply instance segmentation on entire scenes by first applying instance segmentation on each subregion and then integrating the results through a post-processing technique, this approach has a significantly high computational complexity because we need to repeat instance segmentation on each subregion, making it unsuitable for practical use.
Moreover, choosing an appropriate subregion size is difficult, and an integration procedure can add noise to the final result.
Therefore, it is desirable to use as large a region as possible for the input.
However, this figure implies that the instance segmentation task becomes significantly difficult when the input size is large, this difficulty has not been adequately investigated in existing works. Handling large regions, such as an entire scene, is a challenging task; however, it is of great importance in the application of instance segmentation.

\section{Conclusion}
\label{sec:conclusion}
We proposed a new method for instance segmentation on 3D point clouds.
Our memory efficient loss function learns mapping to the embedding space, where the embeddings form clusters for each object. We experimentally showed that our method outperforms existing methods.
Our method can handle a large number of points and performs well even when consuming large regions.
Moreover, we claimed and experimentally demonstrated that existing metrics are not suitable for evaluating instance segmentation because they are considerably affected by the input size of the misclassification.
We proposed novel metrics that are unaffected by such external conditions and can aid in evaluating instance segmentation performances correctly.
Using the proposed metrics, we not only evaluated the instance segmentation task without being affected by external conditions but also analyzed the types of errors in an instance segmentation task for each method.

\section{Acknowledgement}
This work was partially supported by JST CREST Grant Number
JPMJCR1403, and partially supported by JSPS KAKENHI Grant Number
JP19H01115.

{\small

\bibliographystyle{ieee}
}

\end{document}